\documentclass[lettersize,journal]{IEEEtran}
\usepackage{amsmath,amsfonts}
\usepackage{algorithmic}
\usepackage{color}
\usepackage{algorithm}
\usepackage{array}
\usepackage[caption=false,font=normalsize,labelfont=sf,textfont=sf]{subfig}
\usepackage{textcomp}
\usepackage{stfloats}
\usepackage{url}
\usepackage{verbatim}
\usepackage{graphicx}
\usepackage{cite}
\usepackage{multirow}
\usepackage{booktabs}
\begin{document}

\title{Remember and Recall: Associative-Memory-based Trajectory Prediction}

\author{Hang Guo, Yuzhen Zhang, Tianci Gao, Junning Su, Pei Lv$^{\ast}$, and Mingliang Xu
\thanks{Manuscript received xxxx.}
\thanks{$^{\ast}$Corresponding author.}
\thanks{Y. Zhang, P. Lv, and M. Xu, are with the School of Computer and ArtificialIntelligent, Zhengzhou University, Zhengzhou, China. E-mail: {(ielvpei, exumingliang)}@zzu.edu.cn, and {zyzzhang}@gs.zzu.edu.cn.

H. Guo, T. Gao, and J. Su is with the Henan Institute of Advanced Technology, Zhengzhou University, Zhengzhou, China. E-mail: {(guohang, gaotianci, junningsu)}@gs.zzu.edu.cn.}}

\markboth{ieee transactions on intelligent transportation systems}%
{Shell \MakeLowercase{\textit{et al.}}: A Sample Article Using IEEEtran.cls for IEEE Journals}


\maketitle

\begin{abstract}
Trajectory prediction is a pivotal component of autonomous driving systems, enabling the application of accumulated movement experience to current scenarios. Although most existing methods concentrate on learning continuous representations to gain valuable experience, they often suffer from computational inefficiencies and struggle with unfamiliar situations. To address this issue, we propose the Fragmented-Memory-based Trajectory Prediction (FMTP) model, inspired by the remarkable learning capabilities of humans, particularly their ability to leverage accumulated experience and recall relevant memories in unfamiliar situations. The FMTP model employs discrete representations to enhance computational efficiency by reducing information redundancy while maintaining the flexibility to utilize past experiences. Specifically, we design a learnable memory array by consolidating continuous trajectory representations from the training set using defined quantization operations during the training phase. This approach further eliminates redundant information while preserving essential features in discrete form. Additionally, we develop an advanced reasoning engine based on language models to deeply learn the associative rules among these discrete representations. Our method has been evaluated on various public datasets, including ETH-UCY, inD, SDD, nuScenes, Waymo, and VTL-TP. The extensive experimental results demonstrate that our approach achieves significant performance and extracts more valuable experience from past trajectories to inform the current state.

\end{abstract}

\begin{IEEEkeywords}
Trajectory prediction, autonomous vehicles, memory networks, human cognitive process, language models
\end{IEEEkeywords}

\section{Introduction}
\IEEEPARstart{T}{rajectory} prediction is an important technology in various fields, such as intelligent transportation systems~\cite{li2023toward}, autonomous vehicles~\cite{xing2019personalized}, robotics navigation~\cite{he2023robust}, and crowd behavior analysis~\cite{wu2006crowd}. Its goal is to predict the future movement paths of moving objects, playing a crucial role in enhancing traffic management efficiency, ensuring road safety, optimizing robot navigation strategies, and understanding crowd dynamics. Accurate trajectory prediction can not only prevent potential collisions and traffic accidents but also improve system responsiveness and decision-making quality. Furthermore, it also  provides scientific support for urban traffic planning and crowd control, thereby protecting people's lives and property while promoting the efficient development of smart cities and intelligent transportation. Consequently, trajectory prediction has attracted increasing attention from researchers in recent years.

\begin{figure}[!t]
  \centering
   \includegraphics[width=1.0\linewidth]{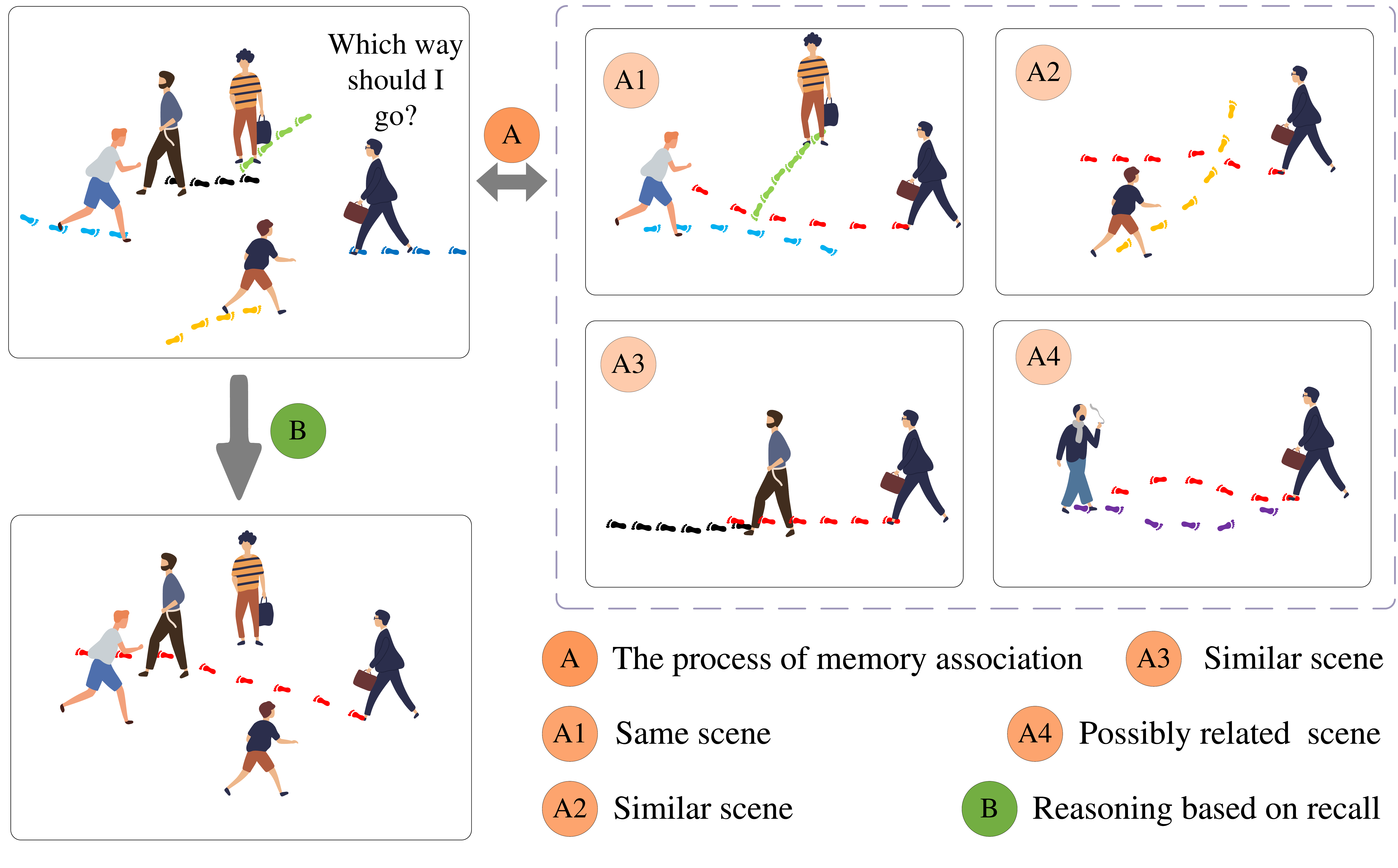}
   \caption{An illustration of the excellent adaptability of humans in dealing with various situations. Facing new environments and challenges, people can quickly extract and analyze fragmented memories from past experiences in order to find experiences that match the current situation. This ability enables humans to process information efficiently and respond rationally and adaptably.}
   \label{figure1}
\end{figure}

To accurately predict the trajectory of moving entities, the key lies in extracting and analyzing the characteristics of motion trajectories from past scenarios, and further enhance the understanding and adaptability to unfamiliar situations. For example, humans can successfully handle a wide variety of tasks, because they have an ability of learning experience, recalling the past events, and generalizing them to an unfamiliar task. As shown in Figure~\ref{figure1}, when pedestrians need to make decisions in complex interactive scenarios, they will quickly search for the associated memory fragments by matching the current scenario with their memory fragments. Then, they further comprehensively extract the key and effective features from these associated memories fragments and even generate some new features based on them, to make the most favorable decision for the current scenario. In this process, they can select the available features closely related to the current scenarios, deeply integrate and analyze these features, and ultimately make the most favorable decision.

To model the cognitive abilities liking human, existing methods usually extract continuous representations from the complete trajectory of an agent as the efficient expression of trajectory. For example, Francesco \textit{et al.}~\cite{marchetti2020mantra} learn continuous representations of trajectories and utilize an associative external memory to store and retrieve these representations. Xu \textit{et al.}~\cite{xu2022remember} store continuous representations of past trajectories to provide a more explicit link between the past trajectory and future. Although significant progress has been achieved, these methods based on continuous representations still exist limitations. First, when handling the continuous representation of trajectory, they face the low computational efficiency due to information redundancy. Second, when applying the learned continuous representations to unfamiliar situations, the continuous representations prevent the model from flexibly matching the suitable representation to the unfamiliar situation, resulting in poor situational adaptability of the model.

To alleviate the above issues, we propose a Fragmented-Memory-based Trajectory Prediction (FMTP) model inspired by the process of learning knowledge, storing information, recalling memory, and applying experience liking human. The core of human driving ability relies on a sophisticated network of neural processes, which is a complex cognitive process~\cite{hajime2001visual}. In FMTP, we design a learnable memory array that participates in the training process of the model to update the stored discrete representations. Subsequently, we exploit a sequential inference engine to reason a combination of discrete representations of future trajectories in the current situation based on the designed memory array. The memory array and inference engine can model the functions of the prefrontal lobe and lateral temporal lobe of human brain, endowing the model with the powerful learning capability and experience application liking humans~\cite{louie2018working,miller2016key}.

The proposed FMTP mainly includes three pivotal elements. Initially, we utilize a joint reconstruction-based architecture composed of two encoders and a decoder to learn rich representations of trajectories. Secondly, to reduce the redundancy of stored information and retain the available features as much as possible, a learnable memory array is designed to fuse continuous representations of trajectories through quantization operations during the training process. Thirdly, in the inference engine based on the language model, we utilize semi-causal intervention model to infer the feature combinations of future trajectories, and further enhance the model's comprehension of discrete representation combination rules and accelerate the convergence speed of model.

\begin{itemize}
    \item We propose a novel FMTP model based on discrete representation, which can significantly enhance computational efficiency and adaptability to different situations by mimicking human learning, memory storage, and experience application processes.
    \item We design a learnable memory array that merges continuous representations of trajectory according to the defined quantization operation. The array can efficiently store and update discrete representations of trajectories, and further eliminate redundant information while maintain the essential features of trajectories.
    \item We develop a Semi-causal intervention inference engine based on language models to deeply learn the associative rule among discrete representations and model the combinatorial properties of trajectory features.
    \item We validate the performance of our method on several public datasets. The extensive experimental results indicate that our approach outperforms existing benchmarks in various metrics, and also reflect the robustness and adaptability on different datasets.
\end{itemize}

\section{Related Work}

\subsection{Trajectory Prediction.}
In the field of trajectory prediction, autonomous driving systems rely on accurate and efficient predictive models for planning and decision-making. Recently, with the significant increases in computing power, deep learning techniques have been widely applied to trajectory prediction and are mainly divided into two types: parameter-based and instance-based methods. Parameter-based frameworks include methods using social pooling~\cite{alahi2016social} and graph neural networks to extract interaction features between agents~\cite{mohamed2020social,huang2019stgat,kosaraju2019social,2024HIMRAE,tang2024hpnet}. Additionally, some studies employ discriminators to score the generated trajectories~\cite{gupta2018social}. For the latter, instance-based frameworks~\cite{marchetti2020mantra,xu2022remember} use an external memory array to store representative instances from the training dataset and search for similar instances from the array during prediction.

Both parameter-based and instance-based methods learn continuous representations through trajectory coordinates. However, continuous representations of different trajectories may contain similar features, leading to redundant information storage in the model and reduced storage efficiency. Additionally, processing the redundant information during the prediction phase easily degrades the performance of model.
In this paper, we propose a discrete representation framework that can effectively capture the key and discrete representations from the agent's trajectory by designing a learning memory array. 
This approach is inspired by the human brain's efficient mechanisms for encoding and storing information, particularly the ability of extracting and generalizing complex information and then fragmentarily storing them as memories.

\subsection{Memory Networks}
Although neural networks with memory capabilities have achieved significant success in multiple fields, they still faces to notable challenges. For example, recurrent neural networks (Recurrent Neural Network, RNN) and their variants~\cite{deo2018multi,alahi2016social,chung2014empirical,2024PoPPL} implicitly store the entire memory state with a state vector, leading to inadequate information processing when handling individual knowledge points independently. Simultaneously, as the amount of information to be remembered gradually accumulates, the model's parameter scale also rapidly grows, which undoubtedly increases computational complexity and resource consumption. To overcome these limitations of RNN, recent research has explored new solutions from Memory-Augmented Neural Networks (MANNs)~\cite{weston2014memory}, which can achieve explicit information storage and precise access to specific features through an externally addressable memory network. MANTR~\cite{marchetti2020mantra} is trained to record and store past and future trajectory pairs, and selectively retain the most valuable samples in a memory array and then gradually build a rich knowledge base to support more accurate predictions. This mechanism cleverly simulates the pattern of human implicit memory to improve the accuracy and generalization of the model. Similarly, MemoNet~\cite{xu2022remember} designs a past and intention memory bank to store features of past-future instance pairs. Through a memory processor, the model can quickly retrieve historical instances highly relevant to the current prediction case from the memory bank. This approach not only optimizes the model's computational efficiency but also establishes a more direct and clear connection between the past and future, enabling the model to make more accurate and coherent predictions.

\subsection{Language models}

Language models are the core concept of natural language processing, and its main function is to understand and generate natural language. The rise of Transformer architecture~\cite{vaswani2017attention} has markedly propelled the evolution of language models in recent years. A distinctive attribute of this architecture is its reliance solely on attention mechanisms for modeling the interplay among inputs. This approach allows the Transformer to accurately process interactions among inputs, regardless of their relative spatial positions. Based on the excellent performance, Transformer architecture has been introduced into the field of trajectory prediction. For example, Yu \textit{et al.}~\cite{yu2020spatio} utilize Transformer to model the spatio-temporal dependencies among trajectories. Roger \textit{et al.}~\cite{girgis2021latent} propose the Transformer-based ``Autobots'' architecture, which alternately performs equivariant processing across the temporal and spatial dimensions. Huang \textit{et al.}~\cite{huang2023gameformer} model the relationships between scene elements by using a Transformer encoder and iterate over the interactions with a hierarchical Transformer decoder structure. Shi \textit{et al.}~\cite{shi2023trajectory} unify trajectory prediction components, social interaction, and multimodal trajectory into an Transformer-based encoder-decoder to effectively eliminate the need for post-processing.


However, in the field of trajectory prediction, most studies merely employ the Transformer as an encoder to extract features from scenes or movement trajectories, which does not fully exploit the Transformer's potential in sequence prediction. Similar to the work~\cite{seff2023motionlm}, our method also directly treats the agent trajectories as motion tokens for trajectory prediction.
The difference is that our method focuses on utilizing the Transformer as a basis to infer future discrete representation sequences and simulate human brain's process of sequentially extracts memories. By learning discrete representations and effectively leveraging the Transformer's sequential prediction capabilities, our framework can flexibly adapt to novel or unfamiliar scenarios. 

\begin{figure*}[t]
  \centering
   \includegraphics[width=0.9\linewidth]{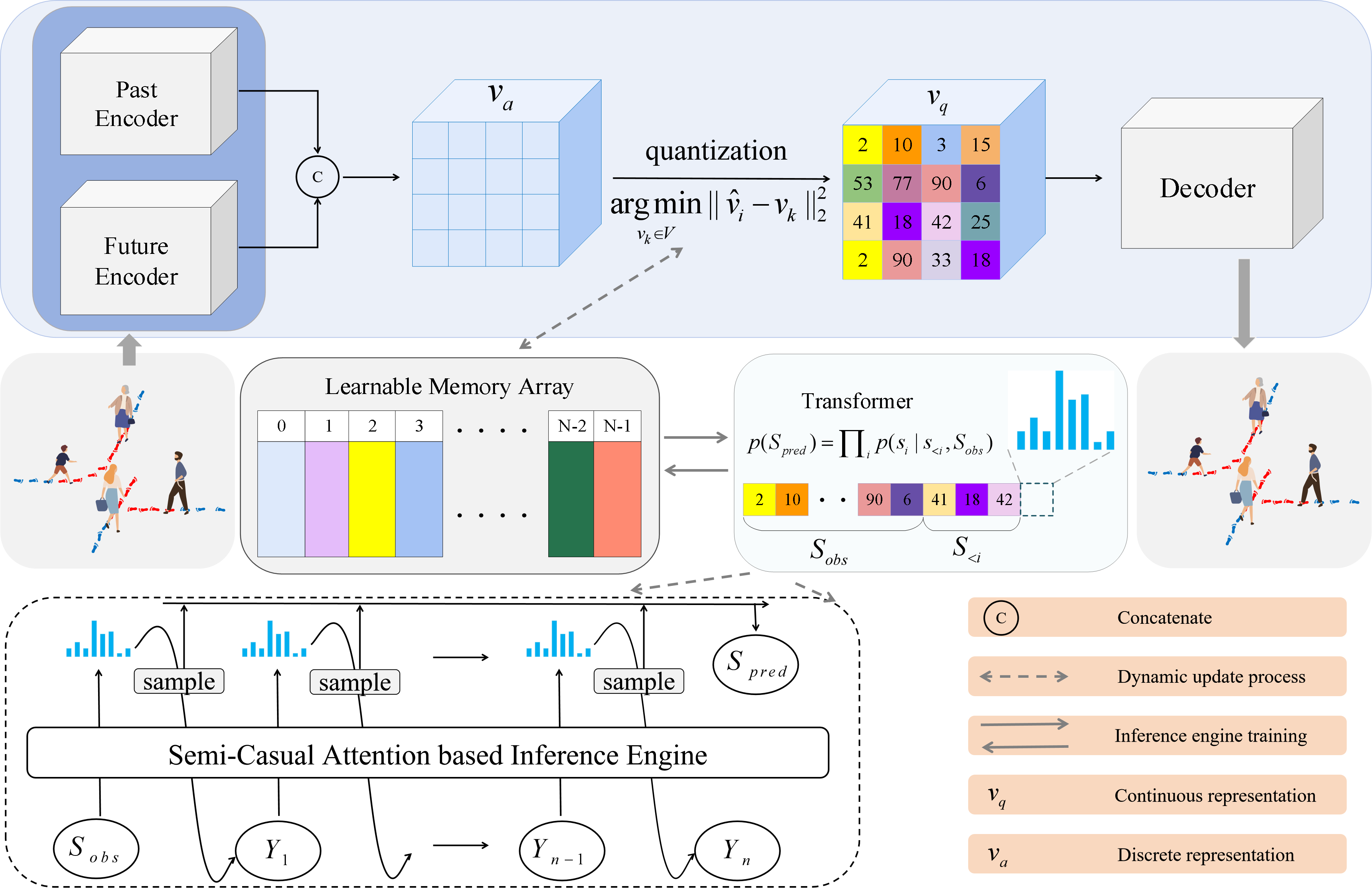}
   \caption{The overall framework of FMTP. In our method, the discrete representations of agent trajectories are captured through quantization operations during the trajectory reconstruction process and are stored in a specially designed memory array. Simultaneously, the encoder and decoder are also trained during this process. We utilize the trained encoder and memory array to extract the index sequences of the trajectory's discrete representations. Based on these index sequences, we employ a Transformer model to accurately model the structure and characteristics of the trajectory's constituent parts.}
   \label{framework}
\end{figure*}

\section{FMTP}
\subsection{FMTP Overview}
The summary and key components of our method are shown in Figure~\ref{framework}, which mainly includes a pair of encoders and decoders, a learnable memory array, and an advanced reasoning engine based on the language models. Therein, the encoder is designed to distill continuous representations from an agent's trajectory. Concurrently, the learnable memory array is used to extract and store discrete representations from these continuous streams. Subsequently, the reasoning engine can adeptly synthesizes these discrete representations. Finally, the decoder interprets the combined discrete representations processed by the reasoning engine to predict the future trajectories of agents.

\subsection{Problem Formulation}

Trajectory prediction involves estimating the future path of an agent based on its past and current states. Mathematically, let $X=\{p_{t}^{i}\}_{t=1}^{T_{obs}}\in\mathbb{R}^{N\times T_{obs} \times2}$ be the past $T_{obs}$ frames of observed trajectory coordinates of N agents, where $p_t^i=(x_t^i,y_t^i)$ represents the 2D
trajectory coordinate of $i$th agent in timestep t.
and let $\hat{Y}=\{\hat{p}_{t}^{i}\}_{t=T_{obs}+1}^{T_{pred}}\in\mathbb{R}^{N\times T_{pred}\times2}$ be the $T_{pred}$ frames of predicted trajectory coordinates continuing from the observed sequences. $T=t_{obs}+t_{pred}$ is the length of entire sequence. $Y$ denotes the ground truth trajectory.

\subsection{Memory Storage Module}
To store discrete representations learned from the trajectories of agents, we design a learnable memory array. The memory array, denoted as $V=\{v_k\}_{k=1}^K\subset\mathbb{R}^{n_k}$, where $n_k$ is the dimensionality of memory array, will provide a rich repository of experiences to bridge the past with the future in the learning process. Consequently, any trajectory $A=\{X||Y\}$ of an agent can be represented as a sequence $V_A=\{v_1,v_2,...v_m\}$ of entries in the memory array. To effectively learn the memory array, we reconstruct a framework based on an encoder $E$ and a decoder $D$ by integrating the concept of neural discrete representation learning. The framework enables the joint learning of how to use the discrete entries $v$ in the memory array to represent the trajectories of agents. More precisely, we employ an encoder to extract continuous features $v_a$ from the trajectory of an agent $i$. Subsequently, each spatial encoding element $\hat{v}_{i}\in{v_a}$ is quantized $q(\cdot)$ to the closest array entry $v_k$, resulting in $v_q$. Finally, we approximate a given agent's trajectory $A$ through $\hat{A}=D(v_q)$.
\begin{equation} \label{eq1}
v_\mathbf{q}=\mathbf{q}(\hat{v})=\left(\underset{v_k\in\mathcal{V}}{\operatorname*{\arg\min}}\lVert\hat{v}_{i}-v_k\rVert\right)\in\mathbb{R}^{m\times n_z}.
\end{equation}

The loss of reconstruction $\hat{A}\approx A$ is then given by
\begin{equation} \label{eq2}
\mathcal{L}_{\mathrm{reconstruction}}=\|A-\hat{A}\|^2.
\end{equation}

In the process of addressing Eq.(\ref{eq2}), we encounter a challenge that these operations involved in the equation are non-differentiable, which precludes the direct use of standard backpropagation algorithms for gradient transmission and updating. To overcome the challenge, we introduce an efficient method for gradient approximation, namely the straight-through gradient estimator. The principle of the estimator is to directly transmit the gradient from the decoder to the encoder~\cite{bengio2013estimating} by bypassing the constraints of non-differentiable operations. This strategy ensures that the model and its memory array can undergo end-to-end training through the loss function as follow:
\begin{equation} \label{eq3}
\mathcal{L}_{\mathrm{VQ}}=\|\mathrm{sg}[E(A)]-v_{\mathbf{q}}\|_2^2\\+\beta\|\mathrm{sg}[v_{\mathbf{q}}]-E(A)\|_2^2,
\end{equation}
where $sg[\cdot]$ denotes the stop-gradient operation, and $\|\mathrm{sg}[v_{\mathbf{q}}]-E(A)\|_2^2$ is the socalled ``commitment loss'' with weighting factor $\beta$~\cite{van2017neural}. In the end, when building the memory array, we also train an encoder capable of extracting trajectory features and a decoder that generates future trajectories based on discrete representational sequences. The comprehensive loss function has been defined as follow:

\begin{equation}
\mathcal{L}(E,G,\mathcal{V})=\mathcal{L}_{\mathrm{reconstruction}}+\mathcal{L}_{\mathrm{VQ}}.
\end{equation}

Through the refined process of reconstruction and vectorization, we can model the advanced mechanisms of the human brain in obtaining, handling, and storing information. Within this framework, we precisely extract distinct data with significant characteristics from observed human motions. This not only simulates the human cognitive process in analyzing and understanding dynamic events, but also enables us to identify and record the most representative and information-rich feature elements in complex datasets, further optimize the efficiency of information storage and retrieval.
\subsection{Memory Association Module}

Once the encoder E, decoder D, and memory array are successfully constructed, we can represent trajectories using the sequence of indices in the memory array constructed by the encoded discrete representations. Consequently, the task of trajectory prediction can be redefined as a Sequence-to-Sequence (Seq2Seq) task~\cite{cho2014learning}: given the discrete representation index sequence $\boldsymbol{S}_{obs}=\{s_1, s_2,...,s_o\}$ of past trajectories to predict the future trajectory discrete representation index sequence $\boldsymbol{S}_{pred}=\{s_{o+1}, s_{o+2},...,s_{o+p}\}$, 
where $o$ and $p$ respectively represent the lengths of the index sequences of the past and future trajectory discrete representations. The sequence of indices is obtained by replacing each entry with its corresponding index in the memory array, and defined as follows:
\begin{equation}v_X,v_Y=q(E(X),E(Y)),\end{equation}
\begin{equation}s_i=k\,\,\,\, such\,\,\,\, that\,\,\,\, \nu_i\in(\nu_X,\nu_Y)=\nu_k.\end{equation}

Similar to traditional Seq2Seq tasks, the language model is trained to perform sequence prediction. First, we use the encoder E to process past trajectory and then transform it into a series of discrete indices, which correspond to specific entries in the memory array. This process can capture the key features of the past trajectory and generate the corresponding index sequence $\boldsymbol{S}_{obs}$ of discrete representations. Next, the sequence is fed into a language model designed for processing and interpreting sequence data. By analyzing the existing sequence patterns, the model generates a new index sequence $\boldsymbol{S}_{pred}$ to represent the predicted future trajectory.
\begin{equation} \label{eq7}
p_b(s_{o+1},s_{o+1},...,s_{o+p}\mid S_{obs})=\prod_{i=o+1}^{o+p}p_b(s_i\mid s_{<i},S_{obs}).
\end{equation}

As suggested in Eq.(\ref{eq7}), the predicted sequence $S_{pred}$ can be expressed as a continuous multiplication of probabilities. More precisely, the prediction of future index sequence can be depicted as autoregressive predictions of subsequent indices. We can employ $p(s_i|s_{<i})$ to ascertain the likelihood of the entire future sequence $S_{pred}$, and optimize the model parameters through the maximization of the likelihood function:
\begin{equation}\mathcal{L}_{\text{Transformer}} = \mathbb{E}_{A\sim p(A)}\left[-\log p(S_{pred})\right].\end{equation}


\subsection{Semi-Casual Relationship Learning}
We introduce a semi-causal attention mechanism in FMTP. Specifically, the output of semi-causal self-attention is calculated as follows:
\begin{equation}\mathrm{Attention_{SC}}=\mathrm{Softmax}\left(\frac{QK^T\times mask_{SC}}{\sqrt{d_k}}\right),\end{equation}
where $Q\in\mathbb{R}^{T\times d_k}$ and $K\in\mathbb{R}^{T\times d_k}$ are query and key respectively, while $mask_{SC}$ is the semi-causal mask with $\mathrm{mask}_{i,j}=-\infty\times1(j\geq\frac s2,i<j)+1(j<\frac s2)+1(j<\frac s2,i\geq j)$, where $s$ is the size of mask.
The semi-causal mask ensures that the future sequence of indexes does not affect the current inference operation in reasoning engine phase. This method allows us to provide rich information when inferring the feature combinations of future trajectories and accelerates the convergence speed of the model.

\section{Experimental And Analysis}
Validating the performance of model on real datasets is necessary~\cite{mo2023map,he2023fear,chen2023stochastic,aksjonov2023safety,zhou2024i2t}. We evaluate the feasibility and effectiveness of our method on several public datasets, demonstrating its predictive performance in complex scenarios and its ability of generalization under different settings.


\subsection{Evaluation Datasets} 

\textbf{ETH-UCY.} The dataset comprises five video sequences from four city environments: ETH and HOTEL from the ETH, and UNIV and ZARA from the UCY. The coordinates of pedestrian in these videos is annotated based on real-world coordinates.

\textbf{inD.} The dataset uses drones to record traffic conditions at four different intersections, extracting trajectories for each road user and their type. It is a public dataset that is widely used in trajectory prediction, scenario-based safety verification of autonomous driving systems, and data-driven development of HAD system components, etc. 

\textbf{nuScenes.} The dataset is a large-scale autonomous driving dataset, and contains 3D bounding boxes of 1000 scenes collected in different countries. Each scene lasts 20 seconds and the annotation frequency is 2Hz. It contains a total of 28130 training samples, 6019 validation samples, and 6008 test samples. 

\textbf{Waymo.} The dataset contains 1950 video clips of 20 seconds collected at 10Hz in various geographical locations and conditions. Moreover, these videos include three types of agents: vehicles, pedestrians, and cyclists. 

\textbf{VTPTL.} The dataset contains about 150 minutes of video clips collected at intersections, T-junctions, and roundabouts. It involves different traffic conditions during peak and off-peak hours, and annotates about 4 million bounding boxes.





\subsection{Evaluation Metrics} 
The performance of our method is evaluated on metrics \textit{Average displacement error} (ADE) and \textit{Final displacement error} (FDE). Note that we used the best-of-K validation method as previous work \cite{gupta2018social}, which means that the prediction with the lowest ADE and FDE among n sampled predictions will be used to measure the accuracy of the model. For the i-th sample, we have:
\begin{equation}ADE_i=\frac1{t_p}\sum_{t=t_o+1}^{t_o+t_p}\parallel p_i^t-\hat{p}_i^t\parallel, \end{equation}

\begin{equation}FDE_i=\parallel p_i^{t_o+t_p}-\hat{p}_i^{t_o+t_p}\parallel. \end{equation}

\begin{table*}[t]
\tiny
	\begin{center}
 \caption{Quantitative comparison with existing methods, using the best-of-20 predictions for each trajectory, on ETH-UCY datasets. All these methods are used to predict the trajectory of the future 12 frames based on the previous 8 frames. The evaluation metrics used in this table are ADE and FDE. 
 }
	\resizebox{\linewidth}{!}{
	\begin{tabular}{c|c|c|c|c|c|c}
        
  \toprule
  
		&ETH&	HOTEL&	UNIV&	ZARA1&	ZARA2&	AVG\\
\midrule
		STGAT~\cite{huang2019stgat}&	0.56 / 1.10&	0.27 / 0.50 &	0.32 / 0.66&	0.21 / 0.42&	0.20 / 0.40&	0.31 / 0.62\\
		Social-STGCNN~\cite{mohamed2020social}&	0.64 / 1.11&	0.49 / 0.85&	0.44 / 0.79&	0.34 / 0.53&	0.30 / 0.48&	0.44 / 0.75\\
		CARPe~\cite{E10}(2020)&	0.80 / 1.48&	0.52 / 1.00 &	0.61 / 1.23&	0.42 / 0.84&	0.34 / 0.74&	0.46 / 0.89\\
		PECnet~\cite{E6}& 0.54 / 0.87 & 0.18 / 0.24 & 0.35 / 0.60& 0.22 / 0.39 & 0.17 / 0.30 & 0.29 / 0.48\\
		Trajectron++~\cite{E++}&	0.43 / 0.86&	0.12 / 0.19 &	0.22 / 0.43&	0.17 / 0.32&	\textbf{0.12} / 0.25&	0.21 / 0.41\\
		GTPPO~\cite{E17}&	0.63 / 0.98&	0.19 / 0.30 &	0.35 / 0.60&	0.20 / 0.32&	0.18 / 0.31&	0.31 / 0.50\\
		SGCN~\cite{E16}&	0.52 / 1.03&	0.32 / 0.55 &	0.37 / 0.70&	0.29 / 0.53&	0.25 / 0.45&	0.37 / 0.65\\
		Introvert~\cite{E18}&	0.42 / 0.70&	0.11 / 0.17 &	0.20 / 0.32&	\textbf{0.16 / 0.27}&	0.16 / 0.25&	0.21 / 0.34\\
        Y-Net~\cite{E21}&	0.28 / 0.33&	\textbf{0.10} / 0.14 &	0.24 / 0.41&	0.17 / \textbf{0.27}&	0.13 / 0.22&	0.18 / 0.27\\
        Goal-SAR~\cite{chiara2022goal}&  0.28 / 0.39&	0.12 / 0.17 &	0.25 / 0.43&	0.17 / 0.26&	0.15 / 0.22&	0.19 / 0.29\\
         MemoNet~\cite{xu2022remember}&  0.40 / 0.61&	0.11 / 0.17 &	0.24 / 0.43&	0.18 / 0.32&	0.14 / 0.24&	0.21 / 0.35\\
        TUTR~\cite{shi2023trajectory}&  0.40 / 0.61&	0.11 / 0.18 &	0.23 / 0.43&	0.18 / 0.34&	0.13 / 0.25&	0.21 / 0.36\\
        HST~\cite{salzmann2023robots}&  0.41 / 0.73&	\textbf{0.10} / 0.14 &	0.24 / 0.44&	0.17 / 0.30&	0.14 / 0.24&	0.21 / 0.37\\
        DynGroupNet~\cite{xu2024dynamic}&  0.42 / 0.66&	0.13 / 0.20 &	0.24 / 0.44&	0.19 / 0.34&	0.15 / 0.28&	0.23 / 0.38\\
        SocialCircle~\cite{wong2024socialcircle}&  0.25 / 0.38 &	0.12 / 0.14 &	0.20 / 0.34&	0.18 / 0.29&	0.13 / 0.22&	0.17 / 0.27\\
        MlgtNet~\cite{10592655}&0.24 / 0.36&\textbf{0.10} / 0.14&0.17 / 0.29&0.17 / 0.28&0.13 / 0.22&0.16 / 0.26\\
        \midrule
        \textbf{FMTP} &	\textbf{0.18 / 0.23}&	\textbf{0.10 / 0.13}&	\textbf{0.15 / 0.23}&	0.18 / 0.29&	0.13 / 	\textbf{0.20}&	\textbf{0.15} / \textbf{0.22}\\
        \midrule
		
	\end{tabular}
    }
 
	\label{tabETH}
    \end{center}
	\centering
\end{table*}

\subsection{Comparison of Baseline Models} 
In order to verify the performance of the proposed method (FMTP), we choose the state-of-the-art methods as the baselines. 
\begin{itemize}
    \item \textit{S-LSTM}~\cite{alahi2016social}, \textit{SR-LSTM}~\cite{B6}: The models based on LSTM are suitable for the processing of sequential data, which can extract the time-series information in the trajectory of the intelligent body and then model the deep interaction between the intelligent bodies.
    
    \item \textit{SGCN}~\cite{E16}, \textit{S-TGCNN}~\cite{mohamed2020social}, \textit{CARPe}~\cite{E10}, \textit{GRIP}++~\cite{F10}: These models cleverly take agents as nodes and the interactions between agents as the connecting edges to build a sophisticated graph structure. Furthermore, through the advanced graph convolutional network learning mechanism, the model can capture the spatio-temporal interaction feature between agents, which not only greatly enhances the rationality and acceptability of predicted trajectories but also significantly reduces the potential risk of collisions.
    
    \item \textit{Introvert}~\cite{E18}, \textit{AMENet}~\cite{2021AMENet}, \textit{SCOUT}~\cite{2021SCOUT}, \textit{GTPPO}~\cite{E17}: When extracting the interactive information between agents, they can focus on key information and eliminate interference by introduce the attention mechanism, and can further capture key information more accurately and quickly.
    
    \item \textit{STGAT}~\cite{huang2019stgat}, \textit{S-BiGAT}~\cite{kosaraju2019social}: These models incorporate Graph Attention Networks (GAT) into trajectory prediction to model the interactions among agents. The graph attention layer computes attention coefficients between node pairs, which are then used to weigh the nodes and update their hidden states. This mechanism allows the model to more effectively capture social interactions among agents, and further generate socially acceptable future trajectories.
    
    \item \textit{PECNet}~\cite{E6}, \textit{Y-Net}~\cite{E21}, \textit{G-SAR}~\cite{chiara2022goal}, \textit{Trajectron}~\cite{E++}: These anchor-guided trajectory prediction methods explicitly account for the impact of destinations on pedestrian paths. They incorporate decision-making modules to plan the optimal route, adapt to environmental changes, simulate individual differences and social interactions, and provide probabilistic outputs with real-time updates to enhance prediction accuracy and robustness.

    \item \textit{LaPred}~\cite{kim2021lapred}, \textit{PGP}~\cite{deo2022multimodal}: These lane-aware trajectory prediction models integrate traffic rules and road structures to provide multimodal and probabilistic predictions, and emphasize real-time performance and environmental adaptability.

    \item \textit{GOHOME}~\cite{gilles2022gohome}, \textit{THOMAS}~\cite{gilles2021thomas}, \textit{HST}~\cite{salzmann2023robots}, \textit{FRM}~\cite{park2023leveraging}, \textit{BIP-Tree}~\cite{zhang2023bip}, \textit{D2-TPred}~\cite{zhang2022d2}, \textit{SocilCircle}~\cite{wong2024socialcircle}, \textit{MlgtNet}~\cite{10592655}: These methods incorporate additional constraints, such as human posture, signal lights, and physical rules, and employ techniques liking heat map and tree structure.
    
\end{itemize}

\begin{table*}[!t]
    \tiny
	\begin{center}
  \caption{Quantitative results on inD dataset with four subsets. Similar to methods~\cite{2021Exploring,2021SCOUT}, we follow the same preprocessing strategy in order to make fair comparison. The ADE and FDE are computed by observing 8 frames (3.2 seconds) to predict 12 future frames (4.8 seconds). Average is the mean of prediction error on four subsets. The bold fonts are the best results with the lowest error among predicted 10 possible trajectories for each agent.}
	\resizebox{\linewidth}{!}{

	\begin{tabular}{c|c|c|c|c|c}
        
             \toprule
		&	IntersectionA &	IntersectionB &	IntersectionC &	IntersectionD & AVG\\
  \midrule
		S-LSTM~\cite{alahi2016social}&2.29 / 5.33 & 1.28 / 3.19 & 1.78 / 4.24 & 2.17 / 5.11 & 1.88 / 4.47 \\
		S-GAN~\cite{gupta2018social} & 3.02 / 5.30 & 1.55 / 3.23 & 2.22 / 4.45 & 2.71 / 5.64 & 2.38 / 4.66 \\
		GRIP++~\cite{F10} & 1.65 / 3.65 & 0.94 / 2.06 & 0.59 / 1.41 & 1.94 / 4.46 & 1.28 / 2.88 \\
		AMENet~\cite{2021AMENet} & 1.07 / 2.22 & 0.65 / 1.46 & 0.83 / 1.87 & 0.37 / 0.80 & 0.73 / 1.59 \\
		 DCENet~\cite{2021Exploring} & 0.96 / 2.12 & 0.64 / 1.41 & 0.86 / 1.93 & 0.28 / 0.62 & 0.69 / 1.52 \\
		SCOUT~\cite{2021SCOUT} & 0.67 / 1.55 & 0.48 / 1.08 & 0.30 / 0.69 & 0.40 / 0.83 & 0.46 / 1.03 \\
		BIP-Tree~\cite{zhang2023bip} & 0.65 / 1.35 & 0.33 / 0.71 & 0.31 / 0.65 & 0.31 / 0.62 & 0.40 / 0.83 \\
  \midrule
        \textbf{FMTP} & \textbf{0.34 / 0.65} &	\textbf{0.18 / 0.36} &	\textbf{0.17 / 0.32} & \textbf{0.12 / 0.20} & \textbf{0.20 / 0.38}\\
        \midrule
	\end{tabular}
	}

	\label{tabinD}
    \end{center}
	\centering	
\end{table*}

\begin{table*}[!t]
    \tiny
	\begin{center}
  \caption{Quantitative results of prediction performance on nuScenes traffic datasets. The subscript represents the value of K
 }
 \resizebox{\linewidth}{!}{
		\begin{tabular}{c|c|c|c|c|c|c|c|c}
   \toprule
    Method &  Trajectron++ & LaPred & GOHOME & Autobot & THOMAS & PGP & FRM & FMTP
			 \\
    \midrule
			 ADE$_5$ & 1.88 & 1.47 & 1.42 & 1.37 & 1.33 & 1.27 & 1.18 & \textbf{1.00}
			 \\
			ADE$_{10}$ & 1.51 & 1.12  & 1.15 & 1.03 & 1.04 & 0.94 & 0.88 & \textbf{0.55}
			 \\
    FDE$_1$ & 9.52 & 10.5  & 6.99 & 8.19 & 6.71 & 7.17 & 6.59 & \textbf{3.24}
			 \\
   \midrule
		\end{tabular}
  }
 
 \label{nuscenes}
	\end{center}
\end{table*}

\begin{table*}[!t]
    \tiny
	\begin{center}
 \caption{Quantitative results of prediction performance on Waymo and VTPTL traffic datasets. 
 }
 \resizebox{\linewidth}{!}{
		\begin{tabular}{c|c|c|c|c|c|c}
   \toprule
    Method &  STGAT  & PECnet & WAYMO-Joint & D2-TPred & MotionLM & FMTP
			 \\
    \midrule
			 Waymo & 1.69 / 3.70 &  0.69 / 1.13 & 0.65 / - & 0.85 / 1.89 & 0.55 / \textbf{1.11} & \textbf{0.52} / 1.27
			 \\
			VTPTL & 21.25 / 43.62 & - & -  & 16.90 / 34.55 & - &\textbf{8.54 / 15.08} 
			 \\
   \midrule
		\end{tabular}
  }

	\label{table3}
	\end{center}
\end{table*}

\subsection{Quantitative Results}

\textbf{ETH-UCY.} For a fair comparison, we select some methods that directly generate 20 possible predicted trajectories without requiring additional post-processing. Simultaneously, we also compare with methods that both ignore video acceleration in the ETH and UNIV subsets and re-segment these two subsets. As shown in Table~\ref{tabETH}, our model demonstrates the superior performance on multiple subsets of ETH-UCY, such as ETH, HOTEL, UNIV, and ZARA2. We attribute the performance to our model's capability to more effectively capture and store the movement characteristics of pedestrian and learn reasonable combinations among them, and further obtain more realistic trajectories. Our method achieves state-of-the-art levels in average ADE and FDE, significantly outperforming other models. Compared to MemoNet, which uses instance memory, our model performs better on all subsets with an average performance improvement of 16.7\% on ADE and 37.1\% on FDE.


\textbf{inD.} we summarize the quantitative results measured by ADE and FDE on inD dataset. The dataset is preprocessed following the same strategy as~\cite{2021Exploring,2021SCOUT} to ensure a fair comparison. For each intersection in inD dataset, 1/3 of recordings are kept for testing, 80\% of the remaining recordings are used for training, and 20\% for validation. We compare our method with 7 related methods, and the corresponding experimental results are shown in Table~\ref{tabinD}. The results indicate that our method all achieves the best performance on the four subsets of inD dataset. On average, our method improves performance by at least 50.0\% on ADE and 54.21\% on FDE.


\textbf{nuScenes.} As shown in Table \ref{nuscenes}, we summarize the quantitative results estimated on nuScenes. Compared with the current state-of-the-art model FRM, the prediction accuracy of our method is improved by the 15.25\%/37.50\%/50.83\% on $ADE_5/ADE_{10}/FDE_1$, respectively. It is worth noting that nuScenes collects samples from 1,000 scenes in different countries with different agent types. These experimental results demonstrate the robustness of our FMTP in different scenarios. Additionally, our method can achieve the best performance when predicting 1/5/10 trajectories, and also shows the excellent multimodality of trajectory.



\textbf{Waymo And VTPTL.} In Table~\ref{table3}, we report the quantitative results estimated on Waymo dataset. Similar to the method~\cite{2021LargeSI}, the same validation parameters are set on Waymo dataset by observing 10 frames to predict the next 30 frames. The quantitative results are reported with the lowest error among the 6 generated possible trajectories. Compared with the PECNET that obtains the better performance by introducing the goal guidance, the performance of our method is improved by 16.67\% on ADE$_{6}$. And we obtain the second-best accuracy on FDE$_{6}$. On VTPTL dataset, compared with D2-TPred introducing the constraint of traffic lights on motion behaviors, the prediction accuracy of our method is improved by 49.47\% on ADE and 56.35\% on FDE. This demonstrates that our method can achieve significant performance in other more challenging scenarios.

\subsection{Ablation Studies}

\textbf{The scale of memory array.} In Table~\ref{inD_one2one}, we present the prediction errors when setting different scale $\theta$ value of memory array.
For these experimental results, we can see that when $\theta$ is set too small, the model loses lots of key information while capturing dispersed representations, and further lead to the decrease of prediction performance. Conversely, when $\theta$ is set too large, the model fails to effectively eliminate redundant information in continuous representations, resulting in an influx of irrelevant information that disrupts the prediction process. The analysis indicates that selecting an appropriate $\theta$ is crucial for enhancing the performance of trajectory prediction. Therefore, to improve the model's predictive accuracy, properly adjusting the value of $\theta$ is key to balance information capture and redundancy elimination. Therefore, under all settings of $\theta$, our model prediction error and memory occupied by the storage module are smaller than the MemoNet method, which indicates that our learnable storage array can extract key information from the trajectory into the designed storage and present excellent performance.

\textbf{Effect of Semi-Casual relationship.} The transformer model commonly employs causal attention to learn the interrelationships between contexts. However, the model easily increases the model's complexity and impedes its convergence rate. To investigate the efficacy of semi-causal interventions, we design the corresponding comparative experiment, and the experimental results are reported in Table~\ref{tabsem}. These results indicate that the model utilizing semi-causal interventions exhibits a markedly faster convergence speed, while maintaining a significant performance of prediction.

\begin{figure}[!t]
  \centering
\includegraphics[width=1.0\linewidth]{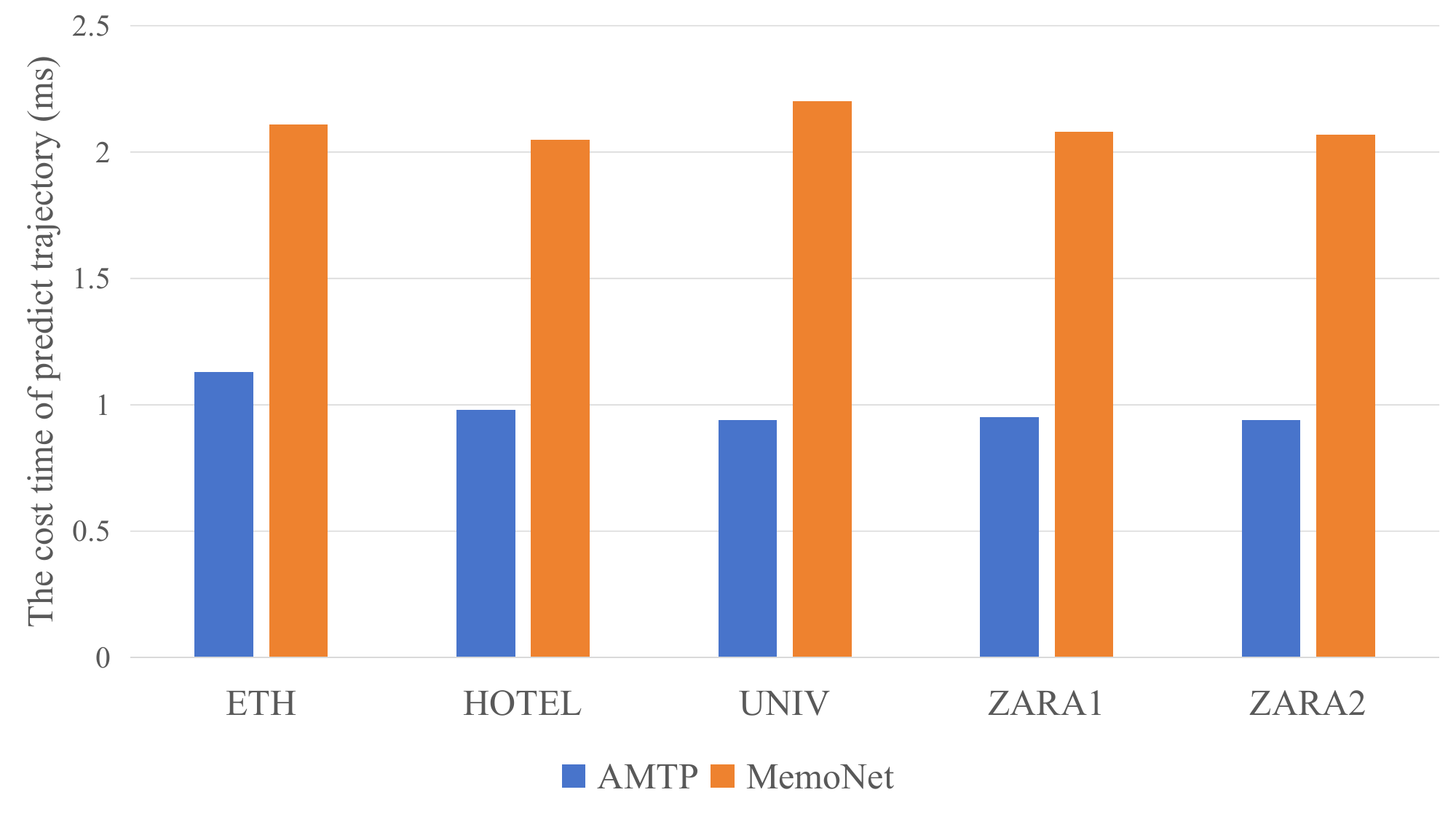}
   \caption{The time cost of FMTP and MemoNet implementing one prediction on the ETH-UCY.}
   \label{time}
\end{figure}

\textbf{Real-time inference speed.} To intuitively demonstrate the efficiency of our method, we visualize the time cost consumption of predicting a single future trajectory for each agent on each subset of the ETH-UCY dataset, as shown in Figure~\ref{time}. These time costs show that our method can complete the prediction of one trajectory within 0.8 to 1.2 milliseconds, which possesses a significant improvement compared with 2.0 to 2.3 milliseconds for MemoNet. It indicates that our method is more computationally efficient and better suitable for implementation in real-world applications.

\begin{table*}[!t]
    \begin{center}
    \tiny
    \caption{Ablation study of entries $\theta$ in memory array on VTPTL. $\theta=768$ achieves the best performance.}
    \resizebox{\linewidth}{!}{
		\begin{tabular}{c|c|c|c|c|c|c}
			\hline
          $\theta$ & MemoNet & 256 & 512 &  768 & 1024 & 1280 
			 \\
    \hline
    Storage & 3.77MB & 17KB & 33KB &  49KB & 65KB & 81KB 
			 \\
			 \hline
			ADE / FDE & 10.11 / 21.37 & 9.76 / 18.75 & 9.36 / 17.25 & \textbf{8.54 / 15.08} & 9.30 / 17.40 & 10.64 / 20.90\\
			 \hline
		\end{tabular}
  }
 \label{inD_one2one}
	\end{center}
\end{table*}

\begin{table*}[!t]
    \tiny
    \begin{center}
    \caption{Ablation study of the impact of model convergence speed using causal and semi-causal relationships. We demonstrate the number of iterations at the time of convergence(IC) on ETY/UCY.}
	\resizebox{\linewidth}{!}{
		\begin{tabular}{c|c|c|c|c|c|c}
   \toprule
           & Metrics & ETH & HOTEL &  UNIV & ZARA1 & ZARA2 
			 \\
    \midrule
			\multirow{2}{*}{Semi-Casual} & IC & 398 & 363 & 543 & 529 & 494 \\
   & ADE / FDE & 0.18 / 0.23 & 0.10 / 0.13 & 0.15 / 0.23&	0.18 / 0.29&	0.13 / 0.20 \\
   \hline
            \multirow{2}{*}{Casual} & IC &	463& 415& 592 & 592&	565 \\
          & ADE / FDE & 0.18 / 0.22 & 0.10 / 0.13 & 0.14 / 0.23&	0.18 / 0.28&	0.13 / 0.20 \\
    \midrule
		\end{tabular}
  }
  
 \label{tabsem}
	\end{center}
\end{table*}

\begin{figure*}[!t]
    \begin{center}
    \includegraphics[width=1.0\linewidth]{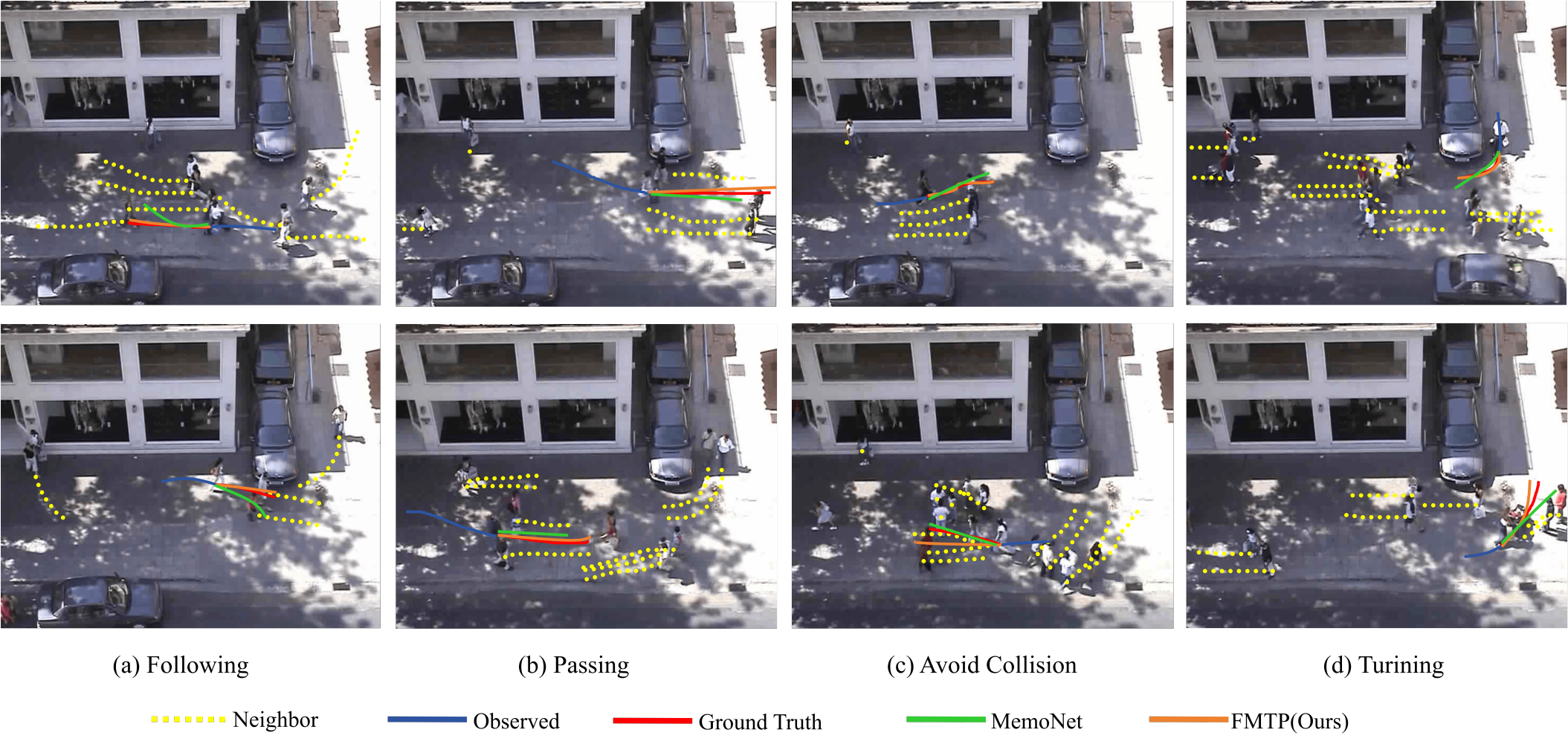}
    \end{center}
    \caption{The visualization of the predicted trajectories by FMTP compared to MemoNet, reproduced with pre-trained weights. Four columns display different motion patterns. To aid in visualization, the trajectories with the best ADE among 20 samples are reported.}
    
    \label{visualization}
 \end{figure*}
 
\begin{figure}[!t]
    \begin{center}
    \includegraphics[width=1.0\linewidth]{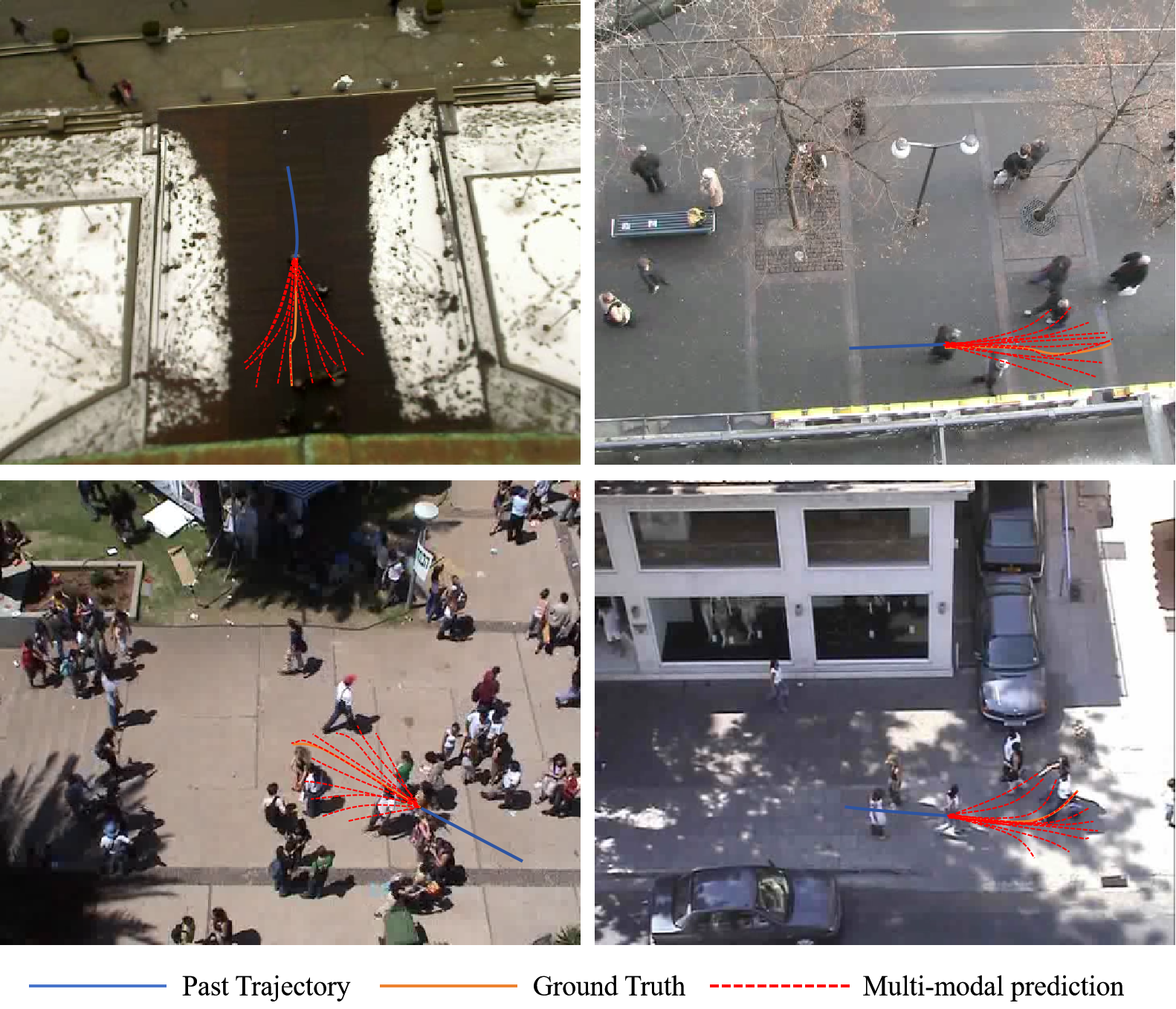}
    \end{center}
    \caption{The visualization results of multimodal prediction for our method on ETH-UCY dataset.}
    
    \label{multi}
 \end{figure}

\subsection{Visualization.}
In addition to the quantitative results, we also carefully select some visual cases of trajectory prediction to qualitatively elucidate the performance of our method. Since the methodology is designed to facilitate a comprehensive understanding of our model in real-world scenarios, the selected visual samples as shown in Figure~\ref{visualization}. 

Figure \ref{visualization} presents a visual comparison of trajectories predicted by various models: Neighbor (yellow), Past path (blue), Ground Truth trajectory (red), MeMoNet (green), and FMTP (orange). These trajectories derive from typical interactive scenarios within the ZARA1 and ZARA2 subsets. Comparing MeMoNet with FMTP, we can see that there are respective strengths and weaknesses between them. In the first column, it depict some scenarios with two pedestrians walking alongside each other or in a leading-following sequence. Comparing with MeMoNet method with conflict and unsmooth trajectory of prediction, our model exhibits a more fluid and natural pattern of movement in these instances. The second column shows cases with a dynamic encounter where a pedestrian is poised to overtake another. MeMoNet exists some shortages in accurately forecasting the impending overtaking maneuver. Conversely, our model can proficiently perceive and predict the nuanced transition. The third column highlights a typical collision avoidance scenario, such as potential collision can be occurred when two pedestrians come together. Our model still expertly generates divergent trajectories to prevent collisions, which emphasizes accuracy of our method in interpreting pedestrian interactions without constructing complex interaction modules.
Finally, the fourth column emphatically demonstrates our model's performance in handling complex situations.
Notably, our model maintains high accuracy of prediction when walking with a large of turning direction, and generates the most appropriate path for pedestrians.

Recent advancements of trajectory prediction methods have emphasized the multimodality of trajectories as a crucial metric for evaluation. To further validate the multimodal capabilities of our model, we carefully select four complex interaction scenarios from the ETH, HOTEL, UNIV, and ZARA subsets, as depicted in Figure~\ref{multi}. These cases show the visual trajectory sample of an individual pedestrian in intricate detail. Our model can not only accurately predict the future trajectory of agents but also generate more socially acceptable predictions. Though not explicitly optimized for multimodal outputs, this capability suggests that our model can effectively recall and process multiple similar or unfamiliar scenarios within the designed learnable storage array. Consequently, it synthesizes agents' responses in these scenarios to generate feasible multimodal trajectories.

\subsection{Limitations.} 
Although the advanced performance has been achieved on multiple public datasets, there are still unreasonable and unexpected failure predictions. Specifically, our model cannot fully capture the complex interactions among pedestrians. For example, in the third column scenarios in Figure~\ref{visualization}, our model can accurately predict collision avoidance with the simple interaction in the first row, but it fails to predict the desired interactive behavior in the complex interaction in the second row. The possible reason is that our model does not include a separate module specifically designed to capture interaction features, which will be the focus of our future work.

\section{Conclusion}
In this paper, we propose a new model FMTP that models the human learning process and their efficient strategy for applying experience. There are two key architectures in our method. One is the discrete representation extraction, which is used to model the classification and conceptualization of information when humans learn. Another is the reasoning engine based on the language models, which simulates how humans use existing knowledge for logical reasoning and decision making. An interesting research for future work may be to incorporate random latent variables into the endpoint prediction to improve the uncertainty of future predictions, such as setting separate latent variables for short-term and long-term goals. Another promising work is to learn the temporal dynamics of the scene to better understand relationships and improve efficiency without restricting the assumption of decomposing attention. Liking most existing work of trajectory prediction, we assume that there is a high-definition map with the marked observation locations of agents, and future work will deploy these solutions to study the impact of imperfect perception on predicting future trajectories.




\bibliographystyle{IEEEtran}
\bibliography{New_IEEEtran_how-to}

\begin{IEEEbiography}[{\includegraphics[
width=1.2in,
height=1.25in,
clip,
keepaspectratio]
{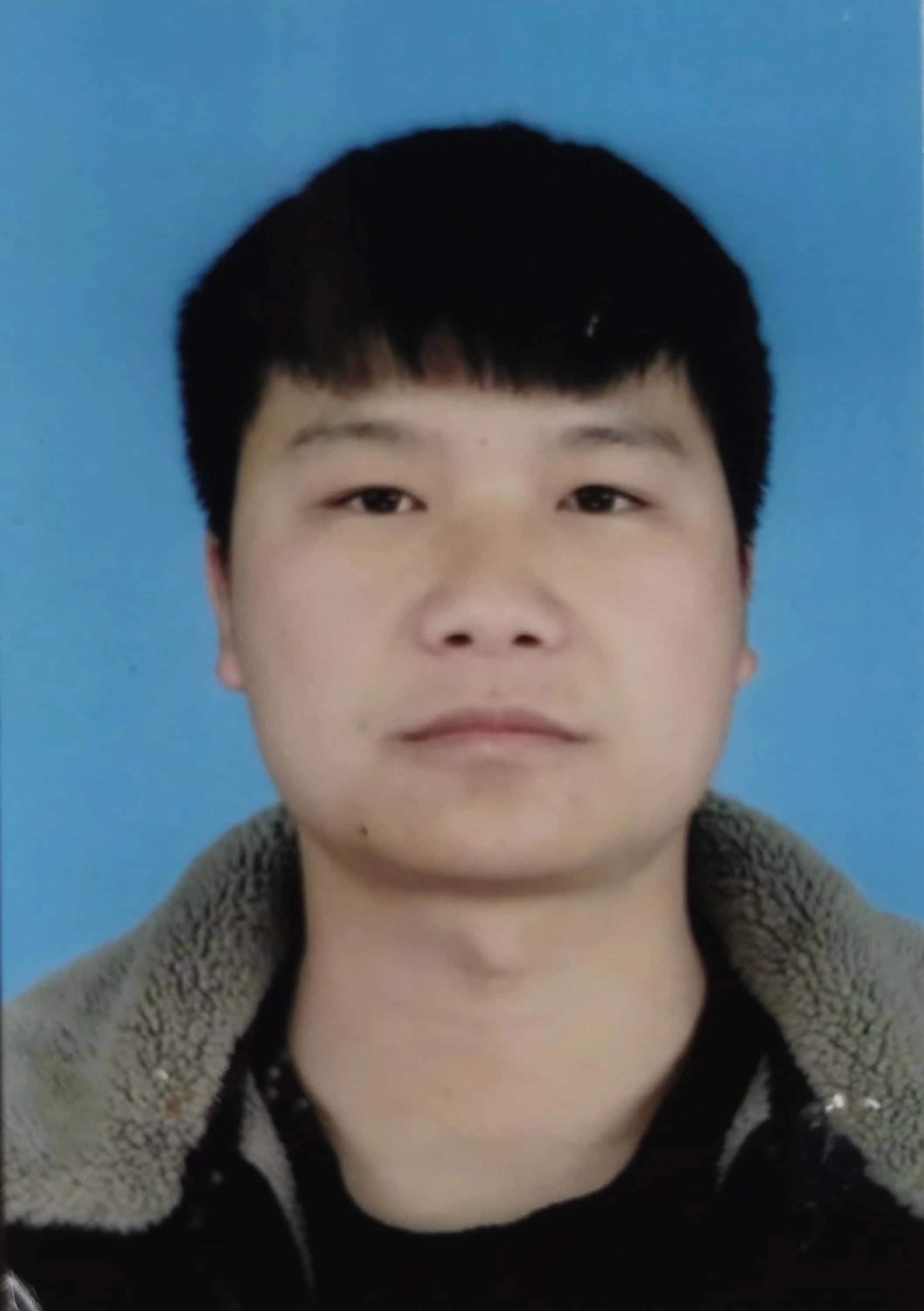}}]
{Hang Guo} received the B.S. degree in Computer Science and Technology from Luoyang Normal University, Guangzhou,China. He is currently pursing the master’s degree in Henan Institute of Advanced Technology, Zhengzhou University, Zhengzhou, China. His current research is focused on deep learning algorithms and applications for autonomous driving, including motion pattern learning for pedestrians and vehicles and future trajectory prediction.
\end{IEEEbiography}
\vspace{-1.1in}

\begin{IEEEbiography}[{\includegraphics[
width=1.2in,
height=1.25in,
clip,
keepaspectratio]
{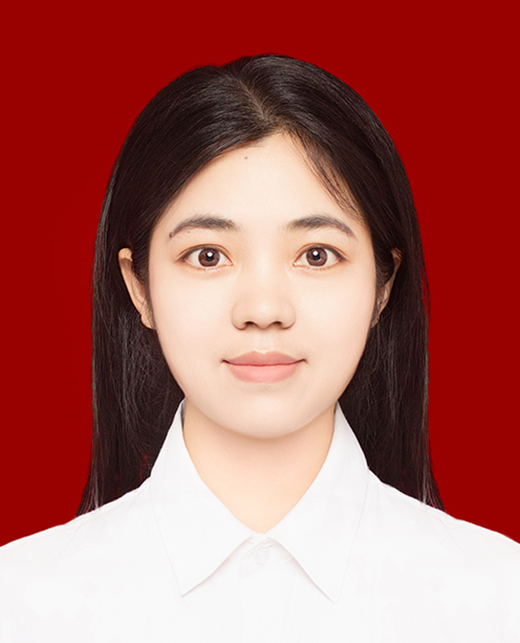}}]
{Yuzhen Zhang} received the B.S. and master’s degrees in software engineering from Henan Polytechnic University, Jiaozuo, China. She is currently pursuing the Ph.D. degree in School of Computer and Artificial Intelligence, Zhengzhou University, Zhengzhou, China. Her current research interests include machine learning, computer vision and their applications to motion prediction, scene understanding, and interaction modeling for intelligent autonomous systems.
\vspace{-1.1in}
\end{IEEEbiography}

\begin{IEEEbiography}[{\includegraphics[
width=1.2in,
height=1.25in,
clip,
keepaspectratio]
{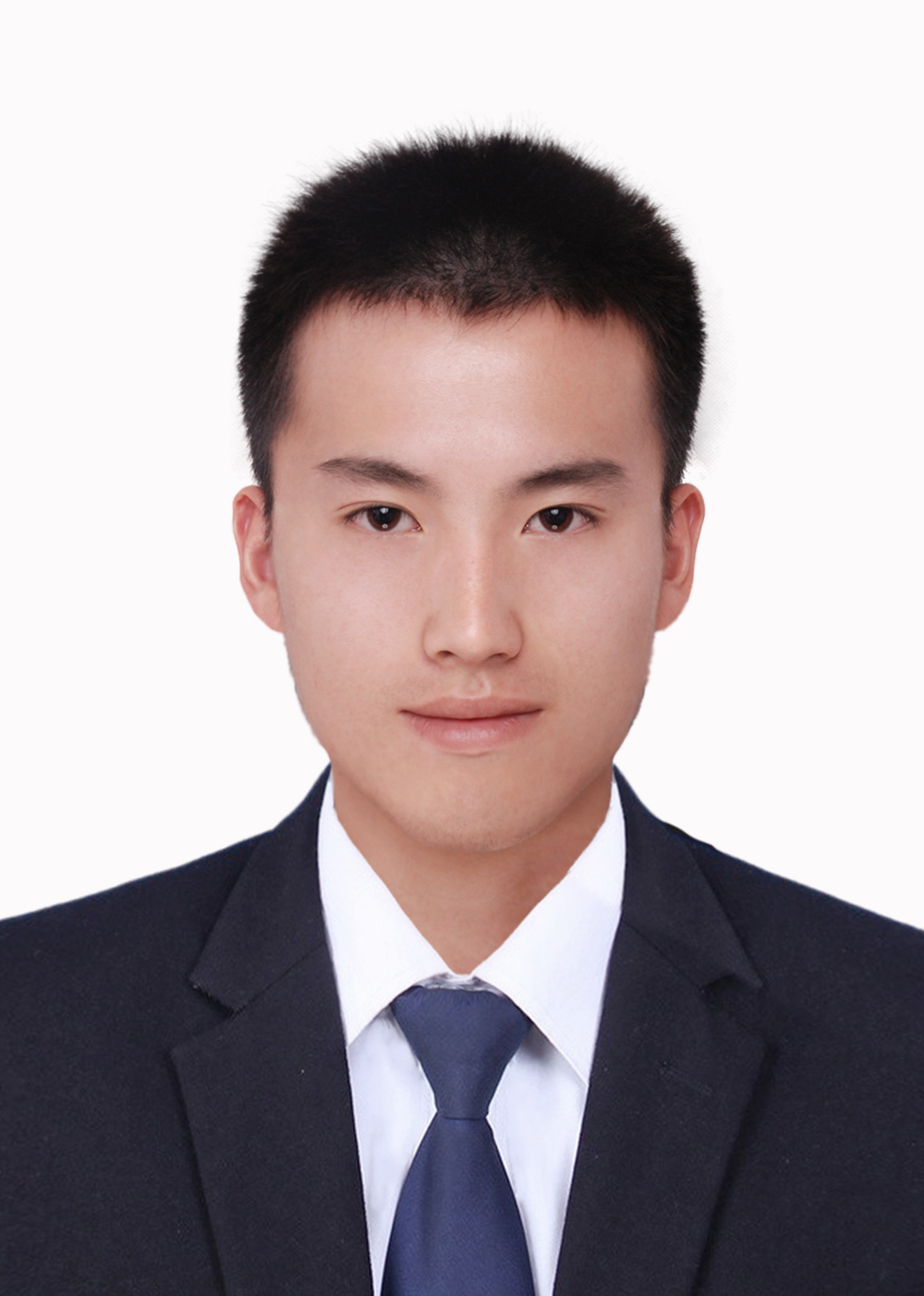}}]
{Tianci Gao} received the B.S. degree in Industry Engineering from Zhengzhou University, Zhengzhou, China. He is currently pursing the master’s degree in Henan Institute of Advanced Technology, Zhengzhou University, Zhengzhou, China. His research primarily focuses on the application of end-to-end learning in the field of autonomous driving, involving areas such as pedestrian trajectory prediction and intent anticipation.
\vspace{-1in}
\end{IEEEbiography}

\begin{IEEEbiography}[{\includegraphics[
width=1.2in,
height=1.25in,
clip,
keepaspectratio]
{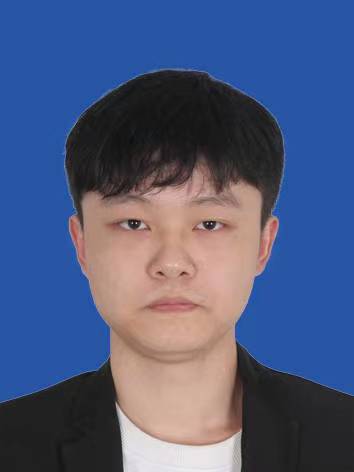}}]
{Junning Su} received the B.S. degree in Animal Science from South China Agricultural University, Guangzhou,China.He is currently pursing the master’s degree in Henan Institute of Advanced Technology, Zhengzhou University, Zhengzhou, China. His current research is focused on deep learning algorithms and applications for autonomous driving, including driving behavior learning, pedestrian trajectory prediction and intention prediction.
\end{IEEEbiography}
\vspace{-1in}

\begin{IEEEbiography}[{\includegraphics[
width=1.2in,
height=1.25in,
clip,
keepaspectratio]
{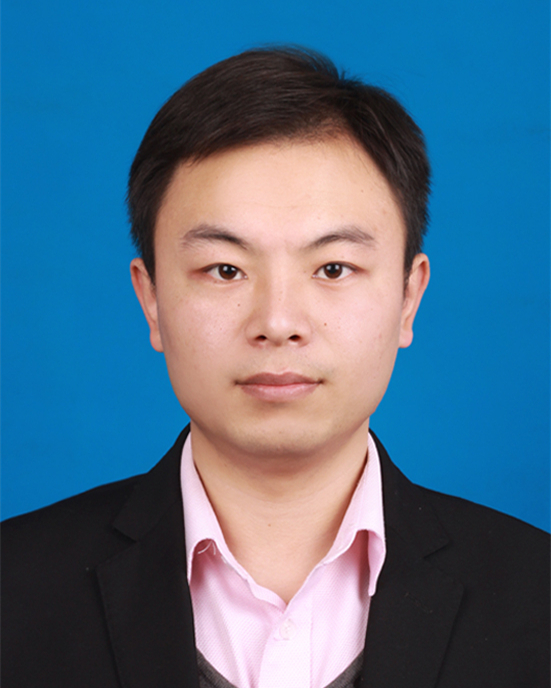}}]
{Pei Lv} received the Ph.D. degree from the State Key Laboratory of CAD\&CG, Zhejiang University, Hangzhou, China, in 2013. He is currently a Full Professor with the School of Computer and Artificial Intelligence, Zhengzhou University, Zhengzhou, China. His research interests include computer vision and computer graphics. He has authored more than 60 journals and conference papers in the above areas, including the IEEE TRANSACTIONS ON IMAGE PROCESSING, the IEEE TRANSACTIONS ON VISUALIZATION AND COMPUTER GRAPHICS, the IEEE TRANSACTIONS ON INTELLIGENT TRANSPORTATION SYSTEMS, IEEE TRANSACTIONS ON AFFECTIVE COMPUTING, the IEEE TRANSACTIONS ON MULTIMEDIA, the CVPR, ECCV, ACM MM, and IJCAI.

\vspace{-1in}
\end{IEEEbiography}

\begin{IEEEbiography}[{\includegraphics[
width=1in,
height=1.25in,
clip,
keepaspectratio]
{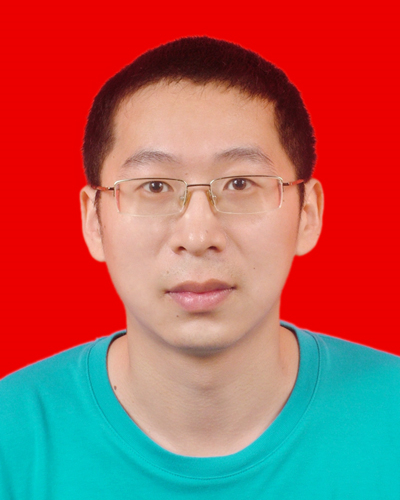}}]
{Mingliang Xu} received the Ph.D. degree in computer science and technology from the State Key Laboratory of CAD\&CG, Zhejiang University, Hangzhou, China, in 2012. He is a Full Professor and the Director with the School of Computer and Artificial Intelligence, Zhengzhou University, Zhengzhou, China. His research interests include computer graphics, multimedia, and artificial intelligence. He has authored more than 100 journal and conference papers in the above areas, including the ACM Transactions on Graphics, the ACM Transactions on Intelligent Systems and Technology, the IEEE TRANSACTIONS ON PATTERN ANALYSIS AND MACHINE INTELLIGENCE , the IEEE TRANSACTIONS ON IMAGE PROCESSING, the IEEE TRANSACTIONS ON CYBERNETICS, the IEEE TRANSACTIONS ON CIRCUITS AND SYSTEMS FOR VIDEO TECHNOLOGY, ACM SIGGRAPH (Asia), ACM MM, and ICCV.
\vspace{-0.6in}
\end{IEEEbiography}

\end{document}